\documentclass[11pt]{article}

\usepackage[preprint]{acl}

\usepackage{times}
\usepackage{latexsym}

\usepackage[T1]{fontenc}

\usepackage[utf8]{inputenc}
\usepackage{algorithm}
\usepackage{algorithmic}
\usepackage{multirow}
\usepackage{amsmath}
\usepackage{amsfonts}
\usepackage{pifont}
\newcommand{\cmark}{\textcolor{green!60!black}{\ding{51}}}
\newcommand{\xmark}{\textcolor{red!70!black}{\ding{55}}}
\usepackage{subcaption}
\usepackage{amssymb}
\usepackage{tcolorbox}
\tcbuselibrary{skins,breakable}

\usepackage{microtype}

\usepackage{inconsolata}

\usepackage{graphicx}

\usepackage{booktabs}
\usepackage[table]{xcolor}

\newtcolorbox{promptbox}[1]{
  colback=gray!5,
  colframe=black!60,
  fonttitle=\bfseries,
  title=#1,
  unbreakable,
  enhanced,
  boxrule=0.5pt,
  arc=1pt,
}

\usepackage{newfloat}
\usepackage{listings}
\DeclareCaptionStyle{ruled}{labelfont=normalfont,labelsep=colon,strut=off} 
\floatstyle{ruled}
\newfloat{listing}{tb}{lst}{}
\floatname{listing}{Listing}
\title{EviDARL: Evidence-Centric Dual-Agent Reasoning via Alternating Reinforcement Learning for Multi-Table Question Answering}

\author{
Dunhan Li$^{1}$, \
Leilei Gan$^{1}$, \
Lirong Gao$^{2}$, \
Wenyan Fan$^{1}$, \\
\bf
Yongwei Wang$^{1}$, \
Sheng Guo$^{2}$, \
Fei Wu$^{1}$ \\
$^1$Zhejiang University,
$^2$MyBank, Ant Group \\
\texttt{nhudl12@zju.edu.cn}
  }

\begin{document}
\maketitle
\begin{abstract}
Despite advances in LLM-based table reasoning, most studies focus on single-table settings. Multi-table QA requires multi-hop reasoning over complex table relationships and evidence dispersed across long contexts, often without verifiable intermediate traces. To address these challenges, we introduce EviDA, an evidence-centric dual-agent framework comprising an Evidence Construction Agent and a Reasoning and Answering Agent. The Evidence Construction Agent generates and executes table-manipulation programs over complete tables to produce question-specific evidence and verifiable intermediate traces, simplifying subsequent reasoning. The Reasoning and Answering Agent then performs multi-hop reasoning over this evidence to derive final answers. We further introduce EviDARL, which combines agent-specific fine-grained rewards with parameter-efficient alternating reinforcement learning to improve collaboration between the agents. By optimizing each agent in turn while holding the other fixed, EviDARL enables bidirectional adaptation between evidence construction and answer reasoning. Experiments on MMQA and TQA-Bench demonstrate that EviDARL substantially outperforms competitive open-model and multi-table reasoning baselines under constrained model and training budgets.

\end{abstract}

\begin{figure}[t]
    \centering
    \includegraphics[width=1.0\linewidth]{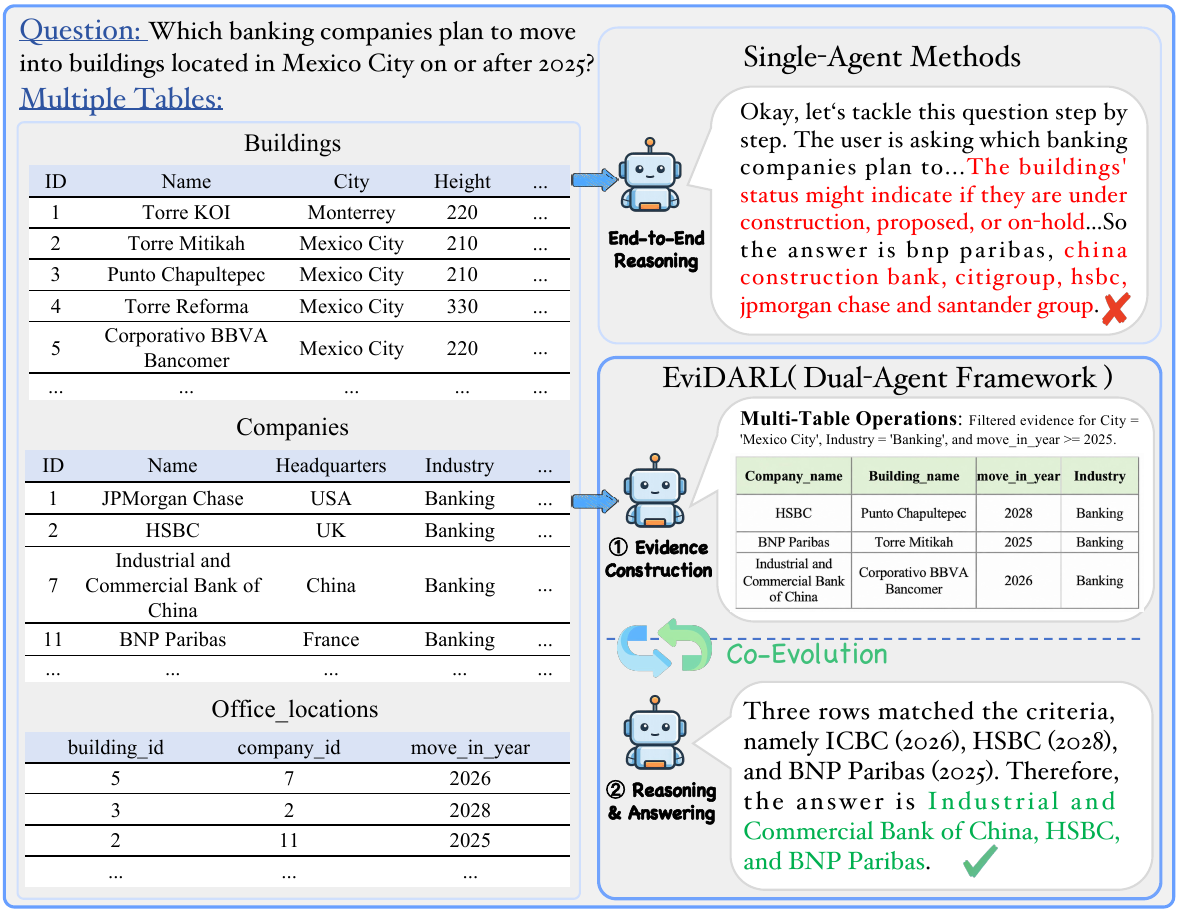}
    \caption{
    An illustrative comparison between EviDARL and single-agent approaches. Single-agent methods may ground reasoning in incorrect information, leading to erroneous reasoning traces and answers, while EviDARL constructs relevant evidence across tables and derives correct answers.
    }
    \label{fig:intro}
\end{figure}

\section{Introduction}

Structured tables are widely used across diverse domains as a compact, information-dense medium for representing, organizing, and communicating complex data. 

Recent work has advanced LLM-based table understanding and reasoning primarily along two directions: post-training LLMs using table-centric data or tasks~~\citep{yang-etal-2025-table, wu-etal-2026-table}, and developing training-free approaches that enhance inference through external components, tailored prompting strategies, or structured reasoning procedures~~\citep{cao2026tablemaster, yu-etal-2025-table, wang2024chainoftable}.

However, existing advances focus mainly on the single table setting.  In real-world applications, structured data are often distributed across multiple relationally connected tables, requiring models to identify cross-table dependencies and integrate evidence from multiple sources. 
This setting, referred to as multi-table question answering (QA), has recently attracted attention from the research community~~\citep{ICLR2025_794a425a,kang-etal-2025-grit,pal-etal-2023-multitabqa,kang-etal-2026-ease}.
Compared with single-table reasoning, multi-table QA poses several additional challenges.
First, multi-table inputs typically introduce substantially longer contexts, in which LLMs often struggle to utilize relevant information robustly, particularly when the necessary evidence is located in the middle of the input or distributed across distant context segments~~\citep{liu-etal-2024-lost, hsieh2024ruler}. 
Second, multi-table QA requires models not only to comprehend the semantics of individual tables, but also to identify relational schemas and to perform operations e.g. cross-table filtering and joining~~\citep{lu-etal-2025-tart}. 
Moreover, the information required for answering a question is often dispersed across multiple tables and must be integrated through multi-step reasoning.  

Several lines of work have begun to address these challenges. 
MultiTabQA serializes the input tables and generates the answer directly. 
However, it is susceptible to long-input truncation and provides no explicit intermediate evidence trace~~\citep{pal-etal-2023-multitabqa}. 
MMQA primarily establishes a benchmark and evaluation framework for multi-table retrieval and reasoning, rather than a trainable pipeline~~\citep{ICLR2025_794a425a}. 
EASE constructs an entity-aware sub-table by extracting query-relevant entities and aggregating selected values from noisy multi-table inputs.
However, its evidence construction and answer generation follow a predefined modular pipeline and are not optimized using downstream answer correctness~~\citep{kang-etal-2026-ease}.
Beyond these studies, reinforcement learning with verifiable rewards offers a natural way to optimize a multi-table reasoning model~~\citep{wu-etal-2026-table}, yet delayed and sparse answer-correctness rewards fail to provide process supervision for multiple intermediate stages, often leading to unstable training.

To bridge the gap, we propose \textbf{EviDA}, an evidence-centric dual-agent framework for multi-table QA that decomposes multi-table reasoning into two complementary stages, as shown in Figure~\ref{fig:intro}. 
In the first stage, an Evidence Construction Agent (EC Agent) receives the question and the multi-table input and generates executable table-manipulation code to construct a compact execution result. The generated program and its execution result jointly form an executable and verifiable evidence trace.
In the second stage, a Reasoning and Answering Agent (RA Agent) performs multi-hop reasoning over the question and the constructed evidence to derive the final answer. The executable evidence alleviates long-context challenges and provides a verifiable interface connecting table operations with reasoning. 

To mitigate the sparse-reward problem of end-to-end RL, we further develop \textbf{EviDARL}, which extends EviDA with fine-grained rewards and parameter-efficient alternating reinforcement learning (RL). 
Specifically, we design fine-grained, role-specific rewards: the EC Agent is rewarded by the downstream utility of both its generated programs and executed evidence, while the RA Agent is optimized primarily with answer-correctness signals. 
Notably, the two policies are bidirectionally dependent: updating the EC Agent shifts the evidence distribution observed by the RA Agent, whereas the RA Agent provides the downstream utility signal for optimizing evidence construction. 
To stabilize this coupled optimization, EviDARL alternately freezes and updates the two agents, so that each policy optimizes against a relatively stable partner while the two progressively co-adapt and co-evolve over multiple rounds of alternation. 
To preserve role-specific policies under a parameter-efficient training and deployment architecture, the two agents share a common backbone while maintaining independent trainable adapters~~\citep{hu2022lora}.

Extensive experiments on MMQA and TQA-Bench validate the effectiveness of both EviDA and EviDARL, demonstrating substantial improvements over competitive baselines. Notably, with only an 8B-parameter model, EviDARL achieves up to 88.40\% accuracy on TQA-Bench and a state-of-the-art exact-match score of 67.80\% on MMQA~~\citep{ICLR2025_794a425a}, surpassing advanced LLMs such as GPT-5.5 and Claude-Sonnet-5 and highlighting both the effectiveness and efficiency of our approach.

\begin{figure*}[!ht]
    \centering
    \includegraphics[width=0.85\textwidth]{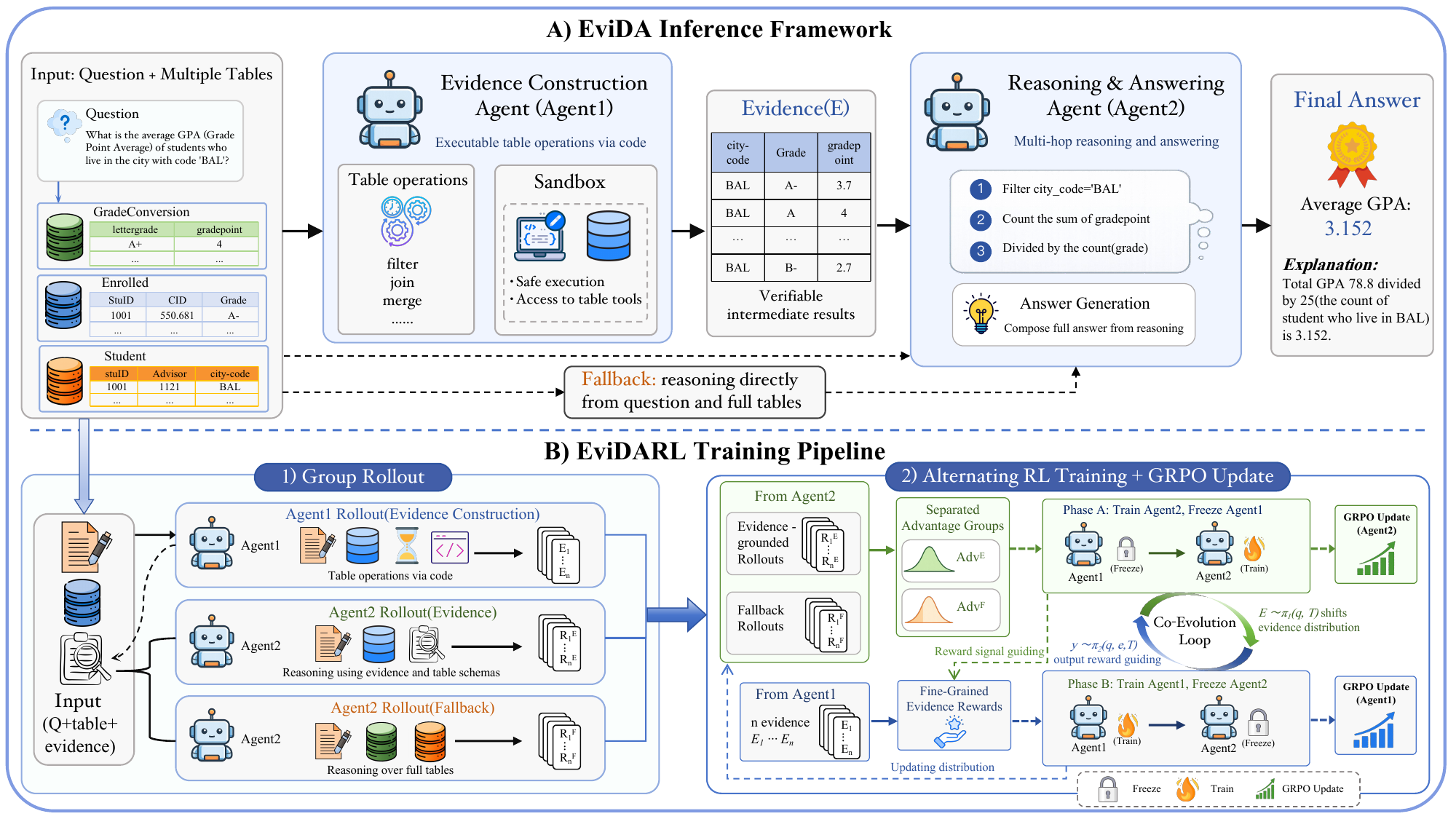}
    \caption{\textbf{Overview of EviDA and EviDARL.} 
    (A) EviDA connects executable evidence construction with evidence-grounded reasoning through a verifiable intermediate interface and a fallback mechanism. 
    (B) EviDARL employs mode-specific rollouts and alternating GRPO updates to enable stable co-evolution between the two agents.}
    \label{fig:overflow}
\end{figure*}

\section{Related Work}

\paragraph{LLMs for Table Reasoning and QA.} 
Recent LLM-based table reasoning methods adapt model parameters to structured table tasks. Instruction-tuned and table-specialized models improve generalization across heterogeneous table operations through table-centric instruction data, synthetic supervision, specialized representations, or large-scale domain adaptation ~\citep{zhang-etal-2024-tablellama,su2024tablegpt2largemultimodalmodel,zhang-etal-2025-tablellm,DBLP:journals/corr/abs-2512-20312}. More recent post-training approaches explicitly optimize multi-step reasoning, reward design and multi-agent~~\citep{yang-etal-2025-table, wu-etal-2026-table, zhou-etal-2026-mixture, yu-etal-2025-table}. The second direction methods augment LLMs with executable tools and structured intermediate representations, using Python or dynamically generated functions for table manipulation and numerical computation, while  progressively transformed table states to support subsequent reasoning~~\citep{lu-etal-2025-tart, wang2024chainoftable, nahid-rafiei-2024-tabsqlify, cao2026tablemaster}. 

However, existing methods are mainly developed for single tables. MultiTabQA provides an early neural formulation for multi-table answer generation ~\citep{pal-etal-2023-multitabqa}, while MMQA~~\citep{ICLR2025_794a425a} systematically evaluates LLMs on multi-table, multi-hop questions requiring an understanding of relational schemas. 
Recent structure-aware methods construct relational context through schema graphs, join-path encoding, or entity-aware sub-table generation ~\citep{kang-etal-2025-grit,wang-etal-2025-plugging,kang-etal-2026-ease}. 
However, they mainly focus on relational representation or evidence reduction before answer generation, while EviDARL jointly optimizes executable evidence construction and evidence-grounded reasoning through alternating RL.

\paragraph{Multi-Agent LLM Reasoning and Reinforcement Learning.}
Early table multi-agent frameworks coordinate frozen LLMs through collaborative criticism and refinement, planning--coding collaboration, and iterative tool-based table exploration ~\citep{yu-etal-2025-table,jin-etal-2025-talon}. These approaches primarily improve inference through predefined communication protocols while leaving the underlying agent policies fixed.

Recent studies move beyond prompted collaboration by explicitly post-training multiple role-specific models~~\citep{park-etal-2025-maporl, wan2025rema, 10.1609/aaai.v40i38.40487}. 
Most closely related to our work, Mixture-of-Minds decomposes table reasoning into planning, coding, and answering roles and optimizes them using RL over MCTS-generated trajectories ~\citep{zhou-etal-2026-mixture}. 
In contrast, EviDARL optimizes evidence construction and answer reasoning with fine-grained, role-specific rewards and alternated optimization. 

\section{Methodology}
In this section, we first introduce \textbf{EviDA}, an evidence-centric dual-agent framework that decomposes multi-table QA into executable evidence construction and evidence-grounded answer reasoning in Sec. \ref{sec:EviMAFramework}. 
We then present \textbf{EviDARL} in Sec. \ref{sec:evimarl-training}, detailing its role-specific reward design, policy optimization objective, and alternating reinforcement learning strategy for collaboratively optimizing the two agents.
\subsection{EviDA Framework}
\label{sec:EviMAFramework}
Multi-table QA requires models to identify question-relevant information distributed across multiple tables, resolve inter-table relationships, and integrate the resulting information to derive the final answer. 
Existing approaches either reason directly over the multi-table input or construct intermediate evidence through predefined or sequential pipelines, the former suffering from long-context interference while the latter fails to jointly optimize evidence construction and answer reasoning.

To address these challenges, we propose \textbf{EviDA}, an evidence-centric dual-agent framework that decomposes multi-table QA into program-based evidence construction and evidence-grounded answer reasoning. 
As illustrated in Figure~\ref{fig:overflow}, EviDA comprises two specialized agents: an \textbf{Evidence Construction Agent} (\textbf{EC Agent}) and a \textbf{Reasoning and Answering Agent} (\textbf{RA Agent}). 
First, the EC Agent generates an executable table-manipulation program from the question and a compact representation of the input tables. 
The program is then executed over the complete table collection to construct question-relevant evidence. 
The resulting evidence is a structured execution result containing filtered records, joined tuples, aggregated values, or other intermediate results required for answering the question. 
Next, the RA Agent reasons over the constructed evidence to derive the final answer. 
If program execution fails or the RA Agent determines that the evidence is insufficient, it switches to fallback mode and reasons directly over the original table collection.
We detail the two specialized agents below.

\subsubsection{Evidence Construction Agent}
\label{sec:ECAgent}
Formally, given a question $q$ and a collection of $n$ tables
$\mathcal{T}= \{T_1,T_2,\ldots,T_n\}$, the EC Agent generates an executable table-manipulation program
$c=\pi_{\mathrm{E}}(q,\mathcal{T}^{\mathrm{cmp}})$
where $\pi_{\mathrm{E}}$ denotes the policy of the EC Agent, $\mathcal{T}^{\mathrm{cmp}}$ is a compact representation of the input tables designed to avoid long-context problems in multi-table QA.
More specifically, we represent each table $T_i$ by its schema and one representative
row, denoted as
$\widetilde{T}_i=
(\operatorname{Schema}(T_i),\operatorname{SampleRow}(T_i))$.
The resulting compact representation is
$\mathcal{T}^{\mathrm{cmp}}
=\{\widetilde{T}_1,\widetilde{T}_2,\ldots,\widetilde{T}_n\}$.
$\mathcal{T}^{\mathrm{cmp}}$ preserves column semantics, representative value formats, and basic structural information while substantially reducing the code-generation context.
Next, the generated code $c$ is executed over the complete table collection $\mathcal{T}$ in a sandboxed environment to produce question-relevant evidence $e=\operatorname{Exec}(c,\mathcal{T})$. 

The generated program and its execution output together form a verifiable evidence trace, allowing evidence-construction errors, such as incorrect joins, inappropriate filters, or missing columns, to be diagnosed separately from downstream reasoning errors.

\subsubsection{Reasoning and Answering Agent}
\label{sec:reasoning-answering-agent}
Given the question-relevant evidence $e$ produced by the EC Agent, the policy of the RA Agent is formulated as \begin{equation} y = \begin{cases} \pi_{\mathrm{R}}^{\mathrm{evi}} \left(q,\mathcal{T}^{\mathrm{cmp}},e\right), & r_{\mathrm{exec}}=1 \land m=1, \\[4pt] \pi_{\mathrm{R}}^{\mathrm{fb}} \left(q,\mathcal{T}\right), & r_{\mathrm{exec}}=0 \lor m=0, \end{cases} \label{eq:ra-policy} \end{equation} where $r_{\mathrm{exec}}\in\{0,1\}$ indicates whether the generated code is executed successfully, and $m\in\{0,1\}$ denotes the evidence-sufficiency decision, with $m=1$ corresponding to sufficient evidence.
This fallback mode ensures robustness: when the generated code fails to execute or the constructed evidence is insufficient, the RA Agent reverts to reasoning directly over the original tables $\mathcal{T}$, so that evidence-construction failures are contained within the framework rather than propagating into prediction failures.

\subsection{EviDARL Training}
\label{sec:evimarl-training}
The two-stage design of EviDA assigns evidence construction and answer reasoning to two specialized agents connected through executable evidence, enabling stage-specific supervision and allowing the two policies to be optimized separately. 
This addresses a key limitation of optimizing both behaviors within a single policy, where sparse final-answer rewards often provide no process-level supervision for either stage.

Building upon the EviDA framework, we introduce EviDARL, which jointly improves the two agents through fine-grained process rewards and an alternating RL strategy. 
Under this strategy, the two agents are optimized in turn. 
As illustrated in Figure~\ref{fig:overflow}, during the optimization of the EC Agent, the RA Agent is frozen and serves as a downstream evaluator of the EC Agent's generated evidence.
Conversely, during the optimization of the RA Agent, the EC Agent is frozen, and the RA Agent is optimized over the evidence it produces.

Next, we detail the reward design for each agent and then introduce the alternating training strategy.\footnote{Note that we first perform supervised fine-tuning (SFT) on both agents using high-quality trajectories distilled from GPT-5.5, providing a stable initialization for subsequent RL.}
\paragraph{Reward Design.}

The EC Agent's reward is gated by code execution: failed execution receives zero reward. For executable code, the reward primarily measures evidence utility through the frozen RA Agent's answer correctness, while also encouraging the retention of question-relevant columns needed for cross-table reasoning. This downstream feedback closes the alternating training loop: EC updates improve the evidence supplied to RA, while the updated RA provides utility feedback for subsequent EC updates. For the RA Agent, we use the final-answer correctness reward
$R_{\mathrm{R}}=r_{\mathrm{ans}}(y,y^*)$, where $y^*$ is the reference answer and $r_{\mathrm{ans}}$ returns $1$ if the predicted answer $y$ matches $y^*$ and $0$ otherwise. The detailed reward formulation is provided in \emph{Appendix}~\ref{sec:reward_detail}.

\paragraph{Optimization Objective.}
We adopt GRPO~~\citep{shao2024deepseekmathpushinglimitsmathematical} as the policy optimization algorithm for both agents. 
For policy $k\in\{\mathrm{E},\mathrm{R}\}$, let
$\{o_i\}_{i=1}^{G}$ be a group of sampled outputs with rewards
$\{R_i^k\}_{i=1}^{G}$. Their normalized advantages are
$\widehat{A}_i^k=(R_i^k-\mu_k)/(\sigma_k+\epsilon_{\mathrm{adv}})$,
where $\mu_k$ and $\sigma_k$ are the group-level reward statistics. 
For the RA Agent, because evidence mode and fallback mode operate on substantially different input distributions, their rewards are grouped and normalized separately rather than compared within the same GRPO group. The GRPO objective is
\begin{equation}
\begin{aligned}
\mathcal{J}_k(\theta_k)
&= \mathbb{E}\Big[
   \frac{1}{G}\sum_{i=1}^{G}
   \frac{1}{|o_i|}\sum_{t=1}^{|o_i|}
   \ell_{i,t}^{k} \\
&\quad - \beta_k D_{\mathrm{KL}}
   \left(\pi_{\theta_k}\|\pi_{\mathrm{ref},k}\right)
   \Big], \\
\ell_{i,t}^{k}
&= \min\Bigl(
   \rho_{i,t}^{k}\widehat{A}_{i}^{k},
   \operatorname{clip}_{\epsilon}(\rho_{i,t}^{k})
   \widehat{A}_{i}^{k}
   \Bigr).
\end{aligned}
\label{eq:grpo-objective}
\end{equation}
where $\rho_{i,t}^{k}$ is the token-level likelihood ratio, $\operatorname{clip}_{\epsilon}(x)
=\operatorname{clip}(x,1-\epsilon,1+\epsilon)$, $\epsilon$ is the clipping coefficient, and $\pi_{\mathrm{ref},k}$ is the SFT-initialized reference policy. The other agent remains frozen during each policy update.

\subsubsection{Alternating Reinforcement Learning Strategy}
\label{sec:alternating-rl}
The two agent policies are inherently coupled: the EC Agent shapes the evidence distribution observed by the RA Agent, while the RA Agent provides the utility reward that supervises evidence construction. 
Updating both simultaneously renders the optimization unstable, as each agent's objective shifts under the other's updates. 
A single pass of sequential training~~\citep{zhou-etal-2026-mixture} instead allows only limited interaction between the two agents.

We address this coupling through the alternating RL. 
Alternating training comprises $K$ cycles, each consisting of two stages: (i) with $\pi_{\mathrm{E}}$ frozen, $\pi_{\mathrm{R}}$ is optimized on evidence sampled from the current evidence policy; and (ii) with the updated $\pi_{\mathrm{R}}$ frozen, $\pi_{\mathrm{E}}$ is optimized using fine-grained rewards, including the downstream utility from the fixed RA Agent. 
The training begins with optimization of the RA policy since the reliability of $r_{\mathrm{util}}$ depends on the RA Agent's ability to evaluate and reason over constructed evidence. 
Through repeated $K$ cycles, the two specialized agents co-evolve progressively: the EC Agent learns to construct evidence that is more useful to the RA Agent, while the RA Agent learns to reason more effectively over the evidence produced by the EC Agent.

\section{Experiments}
\subsection{Experimental Settings}

\paragraph{Datasets.}
We train and evaluate EviDARL on two multi-table QA benchmarks.
(1) \textbf{MMQA}~~\citep{ICLR2025_794a425a} is a challenging benchmark comprising multiple subtasks over relational tables. We focus specifically on its \emph{Multi-Table Multi-Hop QA} subtask. The resulting dataset contains 2,123 training instances and 323 test instances, lightweight data-cleaning as detailed in the \emph{Appendix}~\ref{sec:data_pre}.
(2) \textbf{TQA-Bench}~~\citep{qiu2026tqabenchevaluatingllmsmultitable} evaluates complex reasoning over relational multi-table data in a multiple-choice question (MCQ) format and includes instances at different input-length scales. We sample 1,400 training instances from the 8K and 64K context-length scales, and create two test sets with 500 instances each, one per scale.

\paragraph{Baselines.}
We compare EviDARL against three categories of competitive baselines.
(1) \textbf{Representative advanced LLMs} include Claude Sonnet 5~\citep{anthropic2026sonnet5}, Claude Opus 4.8~\citep{anthropic2026opus48}, DeepSeek-V4-Flash, DeepSeek-V4-Pro~\citep{deepseekai2026deepseekv4}, GLM-5.2~\citep{zai2026glm52}, and GPT-5.5~\citep{openai2026gpt55}.
(2) \textbf{Table-specialized LLMs}: models specifically pretrained or post-trained for table understanding and reasoning, including TableLlama~\citep{zhang-etal-2024-tablellama}, TableGPT2~\citep{su2024tablegpt2largemultimodalmodel}, and TableGPT-R1~\citep{DBLP:journals/corr/abs-2512-20312}.
(3) \textbf{Advanced table-reasoning methods}: representative post-training and multi-agent reasoning approaches that can be extended to multi-table settings, including Table-R1-Zero~\citep{yang-etal-2025-table} and Mixture-of-Minds~\citep{zhou-etal-2026-mixture}.

\paragraph{Metrics.}
We evaluate model performance using five metrics.
(1) \textbf{Exact Match (EM)} .
(2) \textbf{Partial Match (PM)}.
(3) \textbf{Accuracy (Acc.)} is used to evaluate multiple-choice predictions on TQA-Bench.
(4) \textbf{Evidence Construction Success Rate (ECS)} measures the proportion of instances for which the EC Agent successfully constructs evidence.
(5) \textbf{Evidence Sufficiency Rate (ES)} measures, among instances with successfully constructed evidence, the proportion that is factually correct, reasoning-sufficient, and preserves the required cell values, and cross-table relationships.

\paragraph{Implementation details.}
To ensure comprehensive evaluation, we train EviDARL using Llama-3.1-8B-Instruct and Qwen3-8B as backbone models. For all baseline methods, we follow their official implementations and recommended configurations. We apply consistent dataset splits, input representations, and answer-evaluation procedures across methods to ensure a fair comparison. Please refer to \emph{Appendix}~\ref{sec:Exp_detail} for complete implementation details. 

\subsection{Effectiveness of EviDA Framework}
\begin{table}[!t]
    \centering

    \renewcommand{\arraystretch}{1.1}

    \resizebox{\columnwidth}{!}{
    \setlength{\tabcolsep}{2.8pt}
    \begin{tabular}{l @{\extracolsep{4pt}} c @{\hspace{10pt}} c @{\hspace{10pt}} c @{\hspace{10pt}} c c @{\hspace{10pt}} c}
        \toprule
         & \multicolumn{3}{c}{\textbf{MMQA}} & \multicolumn{2}{c}{\textbf{TQA-Bench}} & \\
        \cmidrule(lr){2-4} \cmidrule(lr){5-6}

        \multirow{2}{*}{\textbf{Base Model}} & 2-table & 3-table & Avg & 8K & 64K & \multirow{2}{*}{\textbf{VME}} \\

         & EM & EM & EM  & Acc & Acc & \\

        \midrule
        Claude-Sonnet-5 & 59.28 & 63.73 & 60.69 & 93.40 & 92.00 & \xmark \\
        EviDA(Claude-Sonnet-5) & 68.78\textcolor{green!60!black}{$\uparrow$} & 68.63\textcolor{green!60!black}{$\uparrow$} & 68.73\textcolor{green!60!black}{$\uparrow$} & 93.40\textcolor{gray}{$\rightarrow$} & 92.80\textcolor{green!60!black}{$\uparrow$} & \cmark \\
        \midrule
        Deepseek-V4-Flash & 57.92 & 63.73 & 59.75 & 91.80 & 90.20 & \xmark \\
        EviDA(Deepseek-V4-Flash) & 60.18\textcolor{green!60!black}{$\uparrow$} & 66.67\textcolor{green!60!black}{$\uparrow$} & 62.23\textcolor{green!60!black}{$\uparrow$} & 92.60\textcolor{green!60!black}{$\uparrow$} & 91.40\textcolor{green!60!black}{$\uparrow$} & \cmark \\
        \midrule
        Qwen3-8B & 15.84 & 8.82 & 13.62 & 52.20 & 35.80 & \xmark \\
        EviDA(Qwen3-8B) & 20.36\textcolor{green!60!black}{$\uparrow$} & 20.59\textcolor{green!60!black}{$\uparrow$} & 20.43\textcolor{green!60!black}{$\uparrow$} & 52.80\textcolor{green!60!black}{$\uparrow$} & 40.80\textcolor{green!60!black}{$\uparrow$} & \cmark \\
        \midrule
        Llama-3.1-8B-Instruct & 13.57 & 10.78 & 12.69 & 48.60 & 36.40 & \xmark \\
        EviDA(Llama-3.1-8B-Instruct) & 18.10\textcolor{green!60!black}{$\uparrow$} & 12.74\textcolor{green!60!black}{$\uparrow$} & 16.41\textcolor{green!60!black}{$\uparrow$} & 46.80\textcolor{red!70!black}{$\downarrow$} & 37.60\textcolor{green!60!black}{$\uparrow$} & \cmark \\

        \bottomrule
    \end{tabular}
    }
    \caption{Effectiveness of EviDA across different base models on MMQA and TQA-Bench. VME indicates whether the method constructs verifiable intermediate evidence.} 
    \label{tab:EviMA_framework}
\end{table}

As shown in Table~\ref{tab:EviMA_framework}, EviDA consistently improves MMQA performance across all four base models, yielding average EM gains of 2.48--8.04 points. The largest improvement is achieved with Claude-Sonnet-5, while Qwen3-8B benefits most on three-table questions, increasing from 8.82\% to 20.59\%. On TQA-Bench, EviDA improves six of eight settings, with more consistent gains under the 64K context, indicating that evidence construction facilitates reasoning over long multi-table inputs. Beyond accuracy, EviDA also produces verifiable intermediate evidence, enhancing the transparency of the reasoning process.

\begin{table*}[!t]
    \centering
    \Large
    \renewcommand{\arraystretch}{1.05}
    \setlength{\tabcolsep}{10pt}
    \resizebox{\textwidth}{!}{

    \begin{tabular}{l c  c c  c c c c c c}
        \toprule
         & & \multicolumn{6}{c}{\textbf{MMQA}} & \multicolumn{2}{c}{\textbf{TQA-Bench}} \\
        \cmidrule(lr){3-8} \cmidrule(lr){9-10}

        \multirow{2}{*}{\textbf{Base Model}} & \multirow{2}{*}{\textbf{Method}} & \multicolumn{2}{c}{2-table} & \multicolumn{2}{c}{3-table} & \multicolumn{2}{c}{Avg} & 8K & 64K \\
        \cmidrule(lr){3-4} \cmidrule(lr){5-6} \cmidrule(lr){7-8} \cmidrule(lr){9-9} \cmidrule(lr){10-10}

         & & EM & PM & EM & PM & EM & PM & Acc & Acc \\
        \midrule

        \multicolumn{10}{c}{\textit{\textbf{Proprietary LLMs}}} \\
        \midrule
        Claude-Sonnet-5 & Prompt + Thinking & 59.28 & 71.04 & 63.73 & 73.53 & 60.69 & 71.83 & \underline{93.40} & 92.00 \\
        Claude-Opus-4.8 & Prompt + Thinking & 57.47 & 64.25 & 59.80 & 69.61 & 58.20 & 65.94 & 92.80 & 91.40 \\
        Deepseek-V4-Flash & Prompt + Thinking & 57.92 & 77.38 & 63.73 & 75.49 & 59.75 & 76.78 & 91.80 & 90.20 \\
        Deepseek-V4-Pro & Prompt + Thinking & 61.54 & 77.83 & 66.67 & \underline{79.41} & 63.16 & 78.33 & 92.60 & 91.80 \\
        GLM-5.2 & Prompt + Thinking & 66.52 & 77.38 & \underline{68.63} & 76.47 & \underline{67.19} & 77.09 & 93.20 & \underline{92.20} \\
        GPT-5.5 & Prompt + Thinking & 64.71 & \underline{80.99} & \textbf{70.59} & \textbf{80.39} & 66.56 & \textbf{80.80} & \textbf{94.60} & \textbf{93.40} \\
        \midrule

        \multicolumn{10}{c}{\textit{\textbf{Table-specific LLM}}} \\
        \midrule
        TableLlama-7B & Prompting & 7.69 & 8.60 & 6.86 & 7.84 & 7.43 & 8.36 & OOC & OOC \\
        TableGPT2-7B & Prompting & 17.65 & 23.08 & 11.76 & 19.61 & 15.79 & 21.98 & 52.80 & 36.40 \\
        TableGPT-R1 & Prompt + Thinking & 14.93 & 29.41 & 16.67 & 28.43 & 15.48 & 29.10 & 37.60 & 36.60 \\
        \midrule

        \multicolumn{10}{c}{\textit{\textbf{Multi-table Methods}}} \\
        \midrule
         & Prompting & 13.57 & 18.55 & 10.78 & 14.71 & 12.69 & 17.34 & 48.60 & 36.40 \\
         & Table-R1 & 56.11 & 60.63 & 49.02 & 55.89 & 53.87 & 59.13 & - & - \\
        Llama3.1-8B-Instruct & Mixture-of-Minds & 31.22 & 48.42 & 30.39 & 42.16 & 30.96 & 46.44 & 81.20 & 80.60 \\
         & \cellcolor[HTML]{EFEFEF} \textbf{EviDARL}(ours) & \cellcolor[HTML]{EFEFEF}\underline{67.42} & \cellcolor[HTML]{EFEFEF}73.53 & \cellcolor[HTML]{EFEFEF}53.92 & \cellcolor[HTML]{EFEFEF}63.73 & \cellcolor[HTML]{EFEFEF}63.16 & \cellcolor[HTML]{EFEFEF}70.44 & \cellcolor[HTML]{EFEFEF}87.20 & \cellcolor[HTML]{EFEFEF}86.80 \\
        \midrule
         & Prompting & 15.84 & 21.27 & 8.82 & 14.71 & 13.62 & 19.20 & 52.20 & 35.80 \\
         & Table-R1 & 59.73 & 64.25 & 56.86 & 62.75 & 58.82 & 63.78 & - & - \\
        Qwen3-8B & Mixture-of-Minds & 40.72 & 62.44 & 34.31 & 49.02 & 38.70 & 58.20 & 83.40 & 82.60 \\
         & \cellcolor[HTML]{EFEFEF}\textbf{EviDARL}(ours) & \cellcolor[HTML]{EFEFEF}\textbf{70.59} & \cellcolor[HTML]{EFEFEF}\textbf{81.90} & \cellcolor[HTML]{EFEFEF}61.76 & \cellcolor[HTML]{EFEFEF}71.57 & \cellcolor[HTML]{EFEFEF}\textbf{67.80} & \cellcolor[HTML]{EFEFEF}\underline{78.64} & \cellcolor[HTML]{EFEFEF}88.40 & \cellcolor[HTML]{EFEFEF}88.20 \\
        \bottomrule
    \end{tabular}
    }
    \caption{\textbf{Main results on MMQA and TQA-Bench.} Comparison with proprietary LLMs, table-specific LLMs, and multi-table reasoning methods. The best and second-best results in each column are shown in bold and underlined, respectively.}
    \label{tab:main_results}
\end{table*}

\subsection{Main Results}

\textbf{Comparable Performance with Proprietary and Table-Specific LLMs.}
As shown in Table~\ref{tab:main_results}, EviDARL achieves superior performance across complex multi-table tasks. Although frontier proprietary models such as GPT-5.5 and GLM-5.2 demonstrate substantial reasoning capabilities, achieving average EM scores of 66.56 and 67.19 on MMQA, respectively, EviDARL surpasses both despite using a significantly smaller 8B backbone (Qwen3-8B) and achieves a new state-of-the-art with a 67.80 EM. Furthermore, despite being trained specifically on tabular data, table-specific LLMs such as TableLlama-7B and TableGPT2-7B struggle severely on these multi-table reasoning tasks, yielding low average EM scores of 7.43 and 15.79 due to the complexity of cross-table reasoning. In contrast, EviDARL effectively improves this by an evidence-centric dual-agent learning framework, demonstrating superior accuracy and reasoning capability on these tasks. 

Table 2 also provides a comprehensive quantitative comparison of EviDARL against other multi-table baseline methods under same model architectures. EviDARL significantly enhances the reasoning performance of base LLMs across both the MMQA and TQA-Bench. On Qwen3-8B, EviDARL improves the Average EM from 13.62 (standard prompting) to 67.80 on MMQA, substantially outperforming advanced baselines like Table-R1 (58.82) and Mixture-of-Minds (38.70). A similar improvement is observed with Llama3.1-8B-Instruct, where EviDARL boosts the Average EM to 63.16, compared to 53.87 for Table-R1. EviDARL proves its robustness in handling long-context tabular data by achieving 88.20 accuracy on the Qwen3-8B 64K setup, substantially exceeding existing methods. Additionally, Mixture-of-Minds performs poorly on MMQA, where the likely reason is that its intermediate supervision relies on MCTS-style data generation, which is difficult to scale on complex multi-table, multi-hop questions.

\subsection{Dual-Agent Decomposition and Fine-Grained Process Supervision}
\begin{table}[!ht]
    \centering

    \renewcommand{\arraystretch}{1.1}

    \resizebox{\columnwidth}{!}{
    \setlength{\tabcolsep}{2.8pt}
    \begin{tabular}{l @{\extracolsep{4pt}} c c c @{\hspace{10pt}} c c c}
        \toprule
        \multirow{3}{*}{\textbf{Comparison}} & \multicolumn{6}{c}{\textbf{MMQA}} \\
        \cmidrule(lr){2-7}

         & \multicolumn{3}{c}{2-table} & \multicolumn{3}{c}{3-table} \\
        \cmidrule(lr){2-4} \cmidrule(lr){5-7}

         & EM & ECS & ES & EM & ECS & ES \\
        \midrule
        \addlinespace[2pt]
        EviDARL & \textbf{70.59} & \textbf{95.02} & \textbf{72.38} & \textbf{61.76} & \textbf{88.24} & \textbf{68.89} \\[4pt]
        EviSARL-AC & 57.92 & 75.57 & \underline{62.87} & 48.04 & 73.53 & \underline{60.00} \\[4pt]
        EviSARL-FG & \underline{61.99} & \underline{90.95} & 57.12 & \underline{57.84} & \underline{81.37} & 57.83 \\[4pt]
        Table-R1 & 59.73 & NA & NA & 56.86 & NA & NA \\
        \bottomrule
    \end{tabular}
    }
    \caption{Comparison of dual-agent and single-agent optimization on MMQA.} \label{SAvsMA}
\end{table}

We compare EviDARL with single-agent variants that retain the same evidence-construction and answering workflow but optimize both stages within a unified policy. EviSARL-AC uses only answer correctness and format rewards, whereas EviSARL-FG additionally adopts the fine-grained supervision used in EviDARL; Table-R1 provides an end-to-end single-agent baseline. As shown in Table~\ref{SAvsMA}, EviDARL consistently achieves the best performance across all question scales and evaluation metrics. Compared with EviSARL-AC, it yields EM gains of 12.67 and 13.72 points on two- and three-table questions, respectively, while substantially increasing both ECS and ES. This confirms that delayed answer-level rewards provide insufficient supervision for evidence construction. Introducing fine-grained rewards in EviSARL-FG improves over sparse supervision, particularly raising three-table EM from 48.04\% to 57.84\%. Nevertheless, EviDARL further reaches 61.76\% EM and improves ES from 57.83\% to 68.89\%. Since EviSARL-FG and EviDARL employ comparable signals, this gap shows the benefit of assigning heterogeneous objectives to role-specific policies and alternately optimizing each policy against a stable partner, thereby reducing cross-stage credit-assignment interference. EviDARL also exceeds the end-to-end Table-R1 baseline by 10.86 and 4.9 EM points, further validating the combined effectiveness of dual-agent decomposition and fine-grained process supervision.

\subsection{Ablation Study}
\begin{table*}[!t]
    \centering

    \renewcommand{\arraystretch}{1.1}

    \resizebox{\textwidth}{!}{
    \setlength{\tabcolsep}{2.8pt}
    \begin{tabular}{l @{\extracolsep{4pt}} c c c @{\hspace{4pt}} c c c @{\hspace{4pt}} c c c @{\hspace{10pt}} c c c @{\hspace{4pt}} c c c}
        \toprule
        \multirow{3}{*}{\textbf{Settings}} & \multicolumn{9}{c}{\textbf{MMQA}} & \multicolumn{6}{c}{\textbf{TQA-Bench}} \\
        \cmidrule(lr){2-10} \cmidrule(lr){11-16}

         & \multicolumn{3}{c}{2-table} & \multicolumn{3}{c}{3-table} & \multicolumn{3}{c}{Avg} & \multicolumn{3}{c}{8K} & \multicolumn{3}{c}{64K} \\
        \cmidrule(lr){2-4} \cmidrule(lr){5-7} \cmidrule(lr){8-10} \cmidrule(lr){11-13} \cmidrule(lr){14-16}

         & EM & ECS & ES & EM & ECS & ES & EM & ECS & ES & Acc & ECS & ES & Acc & ECS & ES \\
        \midrule
        Llama3.1-8B-Instruct & 13.57 & NA & NA & 10.78 & NA & NA & 12.69 & NA & NA & 48.60 & NA & NA & 36.40 & NA & NA \\
        EviDA & 18.10 & 72.85 & 4.97 & 12.74 & 60.78 & 3.23 & 16.41 & 69.04 & 4.48 & 46.80 & 67.40 & 21.66 & 37.60 & 63.20 & 23.73 \\
        SFT-EviDA & \underline{54.75} & \underline{79.64} & \underline{54.55} & \underline{48.04} & \underline{71.57} & \textbf{58.90} & \underline{52.63} & \underline{77.09} & \underline{55.82} & \underline{74.20} & \underline{84.60} & \underline{71.16} & \underline{72.80} & \underline{83.40} & \underline{72.42} \\
        SFT-EviDARL & \textbf{67.42} & \textbf{94.57} & \textbf{70.81} & \textbf{53.92} & \textbf{90.20} & \underline{57.61} & \textbf{63.16} & \textbf{93.19} & \textbf{66.78} & \textbf{87.20} & \textbf{98.40} & \textbf{86.79} & \textbf{86.80} & \textbf{97.80} & \textbf{88.55} \\
        \midrule
        Qwen3-8B & 15.84 & NA & NA & 8.82 & NA & NA & 13.62 & NA & NA & 52.20 & NA & NA & 35.80 & NA & NA \\
        EviDA & 20.36 & 82.35 & 4.95 & 20.59 & 68.63 & 17.14 & 20.43 & 78.02 & 8.33 & 52.80 & 82.80 & 20.77 & 40.80 & 80.80 & 24.26 \\
        SFT-EviDA & \underline{58.82} & \underline{85.97} & \underline{62.11} & \underline{53.92} & \underline{77.45} & \underline{64.56} & \underline{57.28} & \underline{83.28} & \underline{62.83} & \underline{79.40} & \underline{89.20} & \underline{82.51} & \underline{76.40} & \underline{88.40} & \underline{81.67} \\
        SFT-EviDARL & \textbf{70.59} & \textbf{95.02} & \textbf{72.38} & \textbf{61.76} & \textbf{88.24} & \textbf{68.89} & \textbf{67.80} & \textbf{92.88} & \textbf{71.33} & \textbf{88.40} & \textbf{99.60} & \textbf{86.14} & \textbf{88.20} & \textbf{99.20} & \textbf{87.10} \\
        \bottomrule
    \end{tabular}
    }
    \caption{\textbf{Ablation of training components on MMQA and TQA-Bench.} EM, ECS, and ES denote exact match, evidence-construction success rate, and evidence sufficie  ncy. Best and second best results are bolded and underlined.} \label{tab:rl_effectiveness}
\end{table*}

As shown in Table~\ref{tab:rl_effectiveness}, SFT provides significant initial gains, while alternating RL consistently yields further improvements across both benchmarks. For Qwen3-8B, SFT-EviDARL increases average MMQA EM from 57.28\% to 67.80\% and TQA-Bench accuracy from 79.40\% to 88.40\% under 8K. Similar gains are observed for Llama3.1-8B-Instruct. The accompanying improvements in ECS and ES indicate that RL enhances both evidence construction and final-answer accuracy.

\subsection{Error Analysis}
To identify where EviDARL succeeds and where further improvements are needed, we analyze errors by failure source and question type.
\begin{figure}[h]
    \centering

    \begin{subfigure}[b]{0.48\columnwidth}
        \centering
        \includegraphics[width=\linewidth]{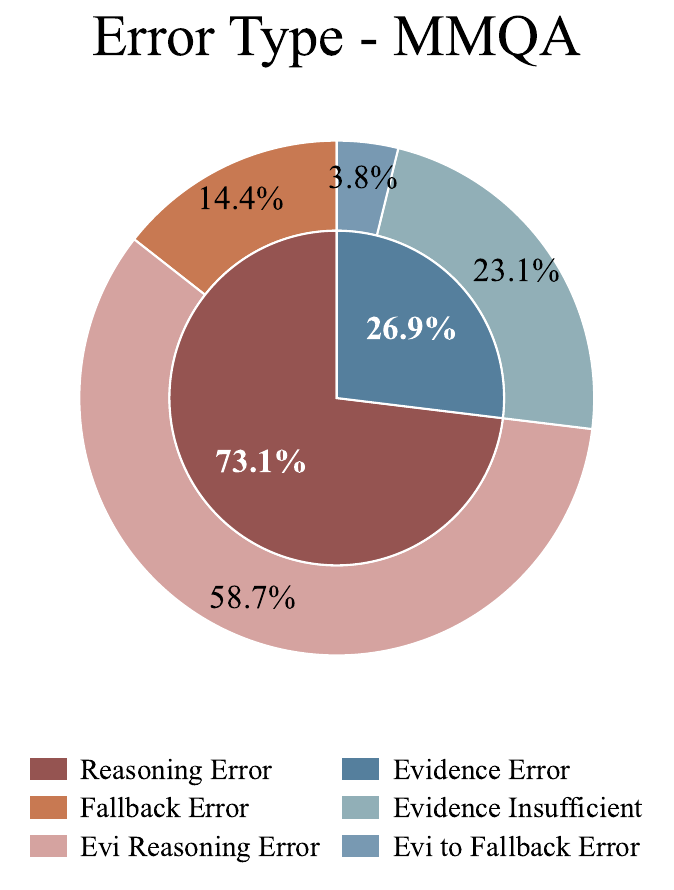}
        \subcaption{}
        \label{fig:case1}
    \end{subfigure}
    \hfill
    \begin{subfigure}[b]{0.48\columnwidth}
        \centering
        \includegraphics[width=\linewidth]{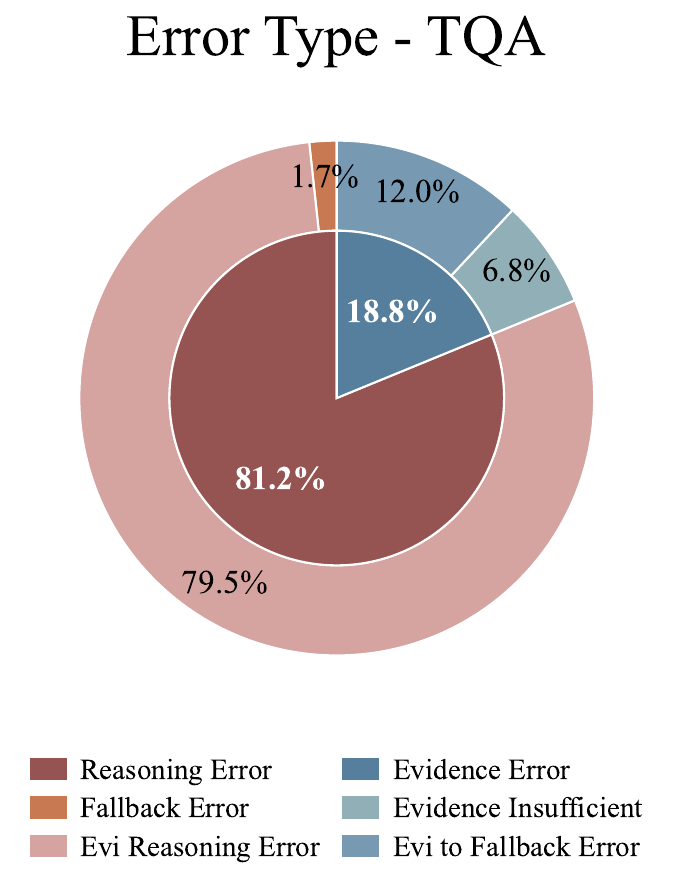}
        \subcaption{}
        \label{fig:case2}
    \end{subfigure}

    \caption{Error distributions of EviDARL on (a) MMQA and (b) TQA-Bench, decomposed into reasoning- and evidence-related failures.}
    \label{fig:error-dist}
\end{figure}
\paragraph{Error Distribution by Failure Source}
As shown in Figure~\ref{fig:error-dist}, reasoning errors dominate on MMQA (73.1\%) and TQA-Bench (81.2\%), particularly in evidence-grounded reasoning, indicating that EviDARL generally constructs usable evidence but remains limited in exploiting it correctly. MMQA shows higher insufficient-evidence and fallback error rates than TQA-Bench (23.1\%/14.4\% vs. 6.8\%/1.7\%), reflecting greater difficulty in evidence construction, whereas TQA-Bench exhibits more evidence-to-fallback routing errors (12.0\% vs. 3.8\%), suggesting imperfect inference-mode selection.

\begin{figure}[t]
    \centering
    \includegraphics[width=1.0\linewidth]{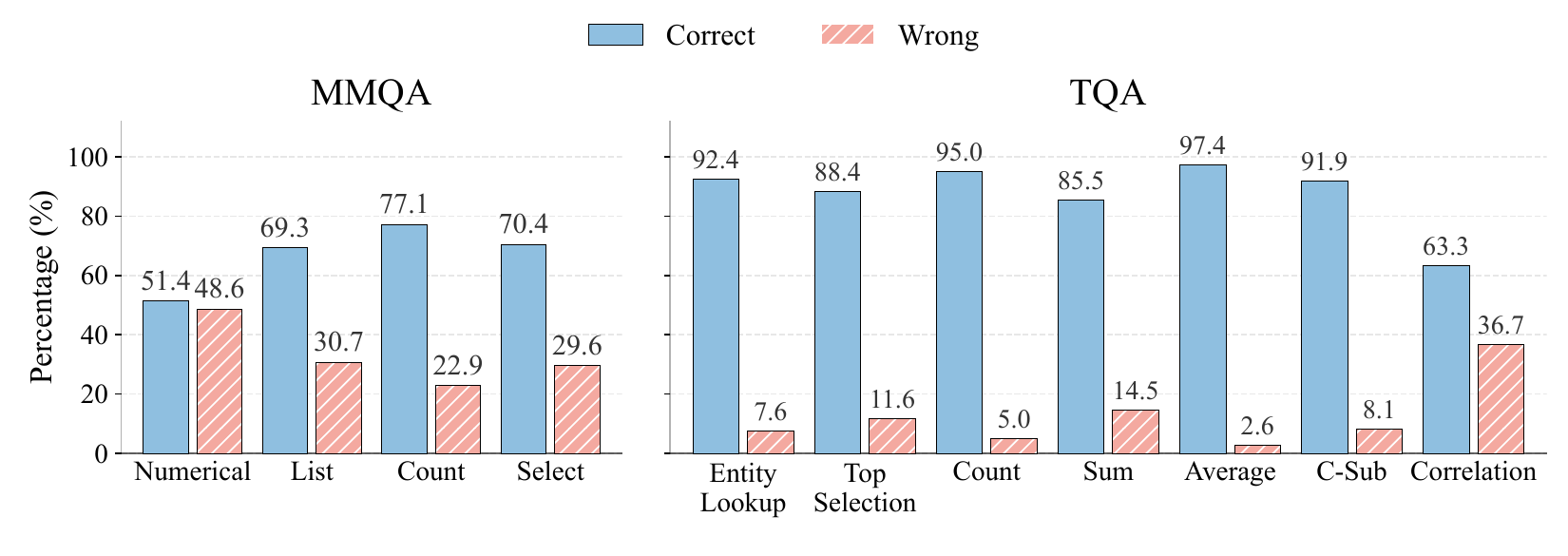}
    \caption{
    Performance of EviDARL on MMQA and TQA-Bench across question types. Bars report the proportions of correct and incorrect predictions within each category.
    }
    \label{fig:MMQA-TQA-quesType-bar}
\end{figure}

\paragraph{Error Distribution of Question Type} Figure~\ref{fig:MMQA-TQA-quesType-bar} further shows strong performance on standard aggregation and entity-level tasks, including Count, Sum, Average, Entity Lookup, and Top Selection, validating the effectiveness of evidence-centric decomposition for structured reasoning. However, numerical questions on MMQA and correlation questions on TQA-Bench remain the most error-prone, while weaker results on List and Select reveal limitations in complex computation and exhaustive cross-table aggregation. 
These findings suggest that future work should strengthen numerical reasoning, evidence completeness, and adaptive routing between evidence-based and fallback inference.

\section{Conclusion}
We propose EviDA, an evidence-centric dual-agent framework that decomposes multi-table QA into executable evidence construction and evidence-grounded reasoning. We further introduce EviDARL, which optimizes the two agents using fine-grained rewards and alternating reinforcement learning. Experiments on MMQA and TQA-Bench show consistent gains over competitive baselines, and ablation studies demonstrate the benefits of dual-agent decomposition, process-level supervision, and alternating optimization.

\section*{Limitations}
Our evaluation covers two benchmarks and two 8B backbones. The pipeline assumes machine-readable tables and does not directly handle table images, charts, or other visual evidence. The RA Agent reasons without calling computation tools, so arithmetic and statistical answers depend on model-generated reasoning; the benefits of tool-augmented answering remain untested.

Executable code provides an inspectable trace, not a guarantee of correct joins, filtering, or complete evidence. LLM-based PM/ES assessments may reflect evaluator bias. Alternation fixes the partner within each stage but does not guarantee convergence, and the experiments do not quantify sensitivity to training randomness. The latency results in \emph{Appendix}~\ref{sec:Latency} characterize the reported hardware and decoding configurations; their generalization to other workloads and deployment settings remains untested.


\bibliography{custom}

\clearpage
\appendix
\section*{Appendix}

\section{EviDA Details}

\subsection{Prompt Details}
We present the detailed prompt templates for the Evidence Construction Agent and the Reasoning and Answering Agent in EviDA, as shown in Figure~\ref{tab:RAprompt}.

\subsection{Compact Representation of Tables}
Directly exposing all cells from the input tables to the agents would reintroduce the context-expansion problem that EviDA is designed to alleviate. We therefore represent each table $T_i$ using only its schema and one representative row, denoted as $\widetilde{T}_i=(\operatorname{Schema}(T_i),\operatorname{SampleRow}(T_i))$, and define the compact multi-table representation as $\mathcal{T}^{\mathrm{cmp}}={\widetilde{T}_1,\widetilde{T}_2,\ldots,\widetilde{T}_n}$. 

The EC Agent primarily needs to understand the schema of each table and the relational schema across tables to identify relevant columns, infer potential primary--foreign key relationships and semantic correspondences, and generate executable table operations. Such structural information is largely conveyed by table and column names. The representative row serves a complementary role by grounding these schema descriptions in concrete values, allowing the agent to infer data types, value formats, identifier patterns, and appropriate operations over heterogeneous fields such as numbers, dates, strings, and entity identifiers. \emph{Since additional rows usually provide little new structural information while substantially increasing context length, a single representative row provides sufficient semantic grounding while preserving effective context compression.} 

The RA Agent likewise receives $\mathcal{T}^{\mathrm{cmp}}$ together with the constructed evidence. While the evidence contains the question-specific records required for reasoning, the compact representation supplements the evidence with schema information, cross-table relations, column semantics, and representative value patterns needed to interpret that evidence correctly. Their combination therefore enables the RA Agent to understand both the provenance and semantic context of the extracted evidence and to derive the final answer without processing the complete tables. The generated program is executed over the full table collection, so compact representation reduces the agents' input contexts without restricting the generated program's access to the underlying tables during execution.

\begin{figure}[!h]
  \centering
  \small

  \begin{promptbox}{Evidence Construction Agent (System)}
You are an expert in table understanding and a Python Code Generator for Data Retrieval.
\#\#\# TASK
Given a user question and table schemas, understand the relations of tables and generate Python code to extract a self-contained and verifiable evidence region that is sufficient for downstream reasoning.
\#\#\# INPUT
- `tables`: A dictionary of DataFrames.
- Table names are available as variables (e.g., if table name is 'Orders', use `Orders` directly).
\#\#\# OUTPUT
- Return ONLY a SINGLE Markdown code block (\verb|```python ... ```|). - The code MUST define a variable `evidence` which is a pandas DataFrame containing the data needed to answer the question. - Handle data types carefully.
  \end{promptbox}

  \begin{promptbox}{Evidence Construction Agent (User)}
Question: \{question\}
Structure of Table:
\{schema-str, SampleRow\}
Generate executable Python code (using pandas) to define `evidence`.
  \end{promptbox}

  \begin{promptbox}{Reasoning and Answering Agent (System)}
You are a professional data analyst in a multi-agent reasoning system.
\#\#\# MODE SPECIFICATION
Each mode defines the quality of evidence available and the reasoning strategy you must apply:

- **Mode1: evidence**
  A code agent successfully extracted a focused evidence DataFrame from the tables.
  Reason strictly from this evidence.
  If the evidence is insufficient (missing columns needed for the logic), set `need-more-evidence` to true.

- **Mode2: fallback**
  Evidence-based reasoning was not possible. You are provided with the complete table data.
  Reason directly from the full tables.
\#\#\# \mbox{Output} format: \mbox{Output} ONE valid JSON object. The `mode` field must exactly match the current mode label.
\{
  "mode": "evidence"|"fallback",
  "reasoning-trace": "Step-by-step logic...",
  "need-more-evidence": boolean,
  "answer": ONLY the exact answer arranged in order, NO other words (string | array | null)
\}
  \end{promptbox}

  \begin{promptbox}{Reasoning and Answering Agent - Evidence (User)}
Mode:
\{mode\}
Question:
\{question\}
Table Schemas:
\{schema-str\}
Row Samples:
\{sample-str\}
Extracted Evidence (DataFrame):
\{evidence-str\}
  \end{promptbox}

  \begin{promptbox}{Reasoning and Answering Agent - Fallback (User)}
Mode:
\{mode\}
Question:
\{question\}
Full-Tables:
\{full-tables\}
  \end{promptbox}

  \caption{System and user prompt templates for EC Agent and RA Agent in EviDA.}
  \label{tab:RAprompt}
\end{figure}

\subsection{SFT Details.}
Using the EviDA prompts described in Figure~\ref{tab:RAprompt}, we distill EC and RA trajectories for all 2,123 training instances.
For EC, we retain trajectories with successfully executed code and assign sample weights of 1.0 and 0.4 when the RA Agent's Evidence Mode answer is correct and incorrect, respectively. This weighting prioritizes evidence that supports correct reasoning while retaining executable code as auxiliary supervision.
For RA, we retain only correctly answered trajectories from both Evidence Mode and Fallback Mode.
As a cold start for alternating RL, SFT establishes basic code generation and answering capabilities, helps reduce execution failures to enable informative utility feedback, and initializes both agents with supervision that favors successful task completion.

\subsection{Reward Details.}\label{sec:reward_detail}
For the Evidence Construction Agent, the reward is defined as
$R_{\mathrm{E}}
=I_{\mathrm{exec}}
(0.20r_{\mathrm{col}}
+0.70r_{\mathrm{util}}
+0.10r_{\mathrm{fmt}}
-0.25p_{\mathrm{full}}
-0.10p_{\mathrm{rep}})$.
Here, $I_{\mathrm{exec}}$ is an execution gate: code that fails to execute receives zero reward, while execution success itself is not treated as an additional reward. 
The column reward $r_{\mathrm{col}}$ assesses whether the constructed evidence retains question-relevant columns, thereby implicitly capturing relational dependencies and semantic correspondences across tables. 
The utility reward $r_{\mathrm{util}}$ measures whether the constructed evidence enables the frozen RA Agent to derive the correct answer, thereby aligning evidence construction with downstream task performance. 
$p_{\mathrm{full}}$ penalizes returning unfiltered or nearly complete tables, whereas $p_{\mathrm{rep}}$ acts as a regularization term that discourages repetitive outputs and prevents the policy from collapsing into degenerate behaviors.
\paragraph{Column Completeness Reward.} For each sample, we use GPT-5.5 to identify the minimal set of columns required to answer the question, using the question and the first ten rows of each table as input. Each column is represented by a table-qualified identifier, i.e., (table name, column name). Let $\mathcal{C}^{*}$ denote this minimal reference column set. The constructed evidence retains a set of columns $\mathcal{C}(e)$. We define a binary completeness signal: $r_{\mathrm{col}} = \mathbb{I}[\mathcal{C}^{*} \subseteq \mathcal{C}(e)]$, which equals 1 if and only if the generated evidence contains all columns in the LLM-provided minimal column list, and 0 otherwise. These required columns implicitly encode both table-level semantics and relational semantics across tables, including semantic correspondences and potential join dependencies. 
\paragraph{Utility Reward.} The constructed evidence is evaluated through the frozen RA Agent in evidence mode, $\hat{y}=\pi_{\mathrm{R}}^{\mathrm{evi}}(q,\mathcal{T}^{\mathrm{cmp}},e)$, and $r_{\mathrm{util}}=\mathbb{I}[\hat{y}=y^{*}]$. This reward receives the largest weight because it directly aligns evidence construction with downstream answer correctness rather than proxy-level evidence properties. 
\paragraph{Repetition Penalty.} To suppress degenerate code generation, we measure code repetitiveness using the zlib compression ratio
$\rho(c)=
|\operatorname{zlib.compress}(c,9)|/|c|$,
where lower ratios indicate stronger verbatim repetition.
The repetition penalty is defined as
\[
p_{\mathrm{rep}}(c)
=
\begin{cases}
0,
& \rho(c) \geq \tau_n, \\[2mm]
\left(
\dfrac{\tau_n-\rho(c)}
{\tau_n-\tau_d}
\right)^{\gamma},
& \tau_d < \rho(c) < \tau_n, \\[3mm]
1,
& \rho(c) \leq \tau_d,
\end{cases}
\]
where $\tau_n=0.3$, $\tau_d=0.1$, and $\gamma=2$. Finally, for highly repetitive outputs with $\rho(c)\leq\tau_{\mathrm{d}}$, the penalty is clamped at $p_{\mathrm{rep}}(c)=1$. This thresholded regularizer leaves normal executable code unpenalized and activates only when the policy degenerates into long repeated spans or collapses toward repetitive outputs.

\section{Experimental Details}\label{sec:Exp_detail}
\subsection{Data Processing Details}\label{sec:data_pre}
We use a fixed random seed of 81 for all data-processing procedures.
\paragraph{MMQA.} For the Multi-Table Multi-Hop QA subset, we first examine samples with empty answer fields using DeepSeek-V4-Flash, which determines whether the question is genuinely unanswerable from the tables and provides a candidate answer when it can be derived; such labels are corrected accordingly. For samples with annotated SQL queries, we execute the SQL and identify discrepancies between the execution results and gold answers. Only samples for which the SQL execution results are inconsistent with the annotated answers are further verified by DeepSeek-V4-Flash, and the gold answers for these cases are corrected based on its judgment. We further remove duplicate instances and a small number of length outliers exceeding 20 times the average sample length in tokens. After preprocessing, the MMQA training portion contains 2,553 instances. We reserve 430 instances for validation and use the remaining 2,123 instances for parameter optimization. The test set contains 323 instances. Thus, the 2,553 training instances reported in the main paper refer to the complete processed training portion before the internal training--validation split. 
\paragraph{TQABench.} The training set contains 1,400 samples across ten domains, with 140 samples drawn from each domain. For each of the 8K and 64K scales, we randomly select database with \texttt{dbIdx} $\in{\{8,9\}}$, randomly sample five \texttt{sampleIdx} values, and include all 14 \texttt{questionIdx} values, yielding $10\times2\times1\times5\times14=1{,}400$ samples. The validation set contains 280 randomly sampled instances from both scales with \texttt{dbIdx} $\in{\{6,7\}}$, unrestricted \texttt{sampleIdx}, and complete coverage of all \texttt{questionIdx} values. We construct separate 8K and 64K test sets of 500 samples each using \texttt{dbIdx} $\in{\{0,\ldots,5\}}$, unrestricted \texttt{sampleIdx}, and complete \texttt{questionIdx} coverage.

\subsection{Metrics Details}
\paragraph{Partial Match(PM).} We follow the partial-match evaluation protocol adopted in MMQA, using DeepSeek-V4-Flash as the evaluator.
\paragraph{Evidence Sufficiency Rate (ES).} To evaluate the quality of the intermediate evidence constructed by the EC Agent, for every instance with successful evidence construction, we use DeepSeek-V4-Flash to assess whether the evidence contains the critical rows, columns, cell values, and cross-table relationships required to answer the question completely. ES is computed as the number of samples satisfying these criteria divided by the total number of successfully constructed evidence samples. As a key metric for the evidence-construction objective, ES measures the quality and sufficiency of the verifiable intermediate evidence produced by EviDA.

\subsection{Baseline Details \& Analysis}
\paragraph{Implementation.} For Mixture-of-Minds, we use the publicly released code and follow its original prompts, training procedure, and reported hyperparameters. For Table-R1, we follow the two DAPO techniques specified in the original work: token-level loss computation and asymmetric decoupled clipping. 
\paragraph{Detailed Analysis.} As reported in the main results, Mixture-of-Minds performs poorly on MMQA, achieving only 40.72\% and 34.31\% EM on two- and three-table questions with Qwen3-8B; a similar trend is observed with Llama-3.1-8B-Instruct. A likely reason is that its intermediate supervision relies on MCTS-style data generation, which is difficult to scale on complex multi-table, multi-hop questions. Among the 2,553 instances in the processed MMQA training portion, comprising 2,123 optimization instances and 430
held-out validation instances, the plan, code, and answer agents
produce only 423, 345, and 579 valid trajectories, respectively; this limited supervision likely contributes substantially to the weaker performance.

\subsection{Training Details}
\paragraph{Hardware Settings.} We train EviDARL on four NVIDIA A100 80GB GPUs with 400GB of system memory and 48 CPU cores. Mixture-of-Minds and Table-R1 are trained on eight NVIDIA A100 80GB GPUs with 800GB of system memory and 96 CPU cores. 
\paragraph{Hyperparameters.}
For GRPO optimization, we use a rollout group size of 8, a batch size of 32, a learning rate of $3 \times 10^{-5}$, and a rollout temperature of 0.8. We train EviDARL for 6 epochs corresponding to 396 steps. The hyperparameter $K$ controls the number of complete alternation cycles(see Section ~\ref{RLtraining} for detailed experiments). We set $K=3$, corresponding to three complete alternation cycles, each comprising one epoch of RA-Agent optimization followed by one epoch of EC-Agent optimization.
\paragraph{Parameter-Efficient Training and Deployment.} Notably, in EviDARL, both agents share the same frozen backbone, and each agent maintains one distinct trainable LoRA adapter. For Qwen3-8B and Llama-3.1-8B-Instruct, each role-specific adapter contains approximately 87.3M and 83.4M trainable parameters, respectively, accounting for about 1\% of the corresponding backbone. Across the two agents, the total adapter parameters are approximately 174.6M and 166.8M, respectively, although only one adapter is optimized at a time during alternating training. This parameter-efficient design significantly reduces optimization cost while preserving strong task-specific adaptation capability across agents. Importantly, compared with Mixture-of-Minds, which instantiates separate full backbones for its three roles, EviDARL uses a single shared backbone with two lightweight role-specific adapters, substantially reducing the model-parameter footprint for training and deployment.

\subsection{Supplementary Analysis}
\paragraph{PM Analysis.} The PM results further confirm the effectiveness of EviDARL beyond strict exact-match accuracy. With Qwen3-8B, EviDARL achieves 81.90 and 71.57 PM on two- and three-table questions, respectively, yielding an average PM of 78.64 that ranks second among all evaluated methods and approaches GPT-5.5 (80.80) despite using a substantially smaller backbone. It also exceeds Table-R1 and Mixture-of-Minds by 14.86 and 20.40 average PM points, respectively, with similar gains observed with Llama-3.1-8B-Instruct. The gap between PM and EM mainly reflects predictions that are semantically or partially correct but fail exact matching because of numerical precision or rounding differences, incomplete enumeration in list-type questions, or minor formatting variations. These results indicate that EviDARL improves not only exact-answer accuracy but also the semantic correctness and information coverage of predictions that do not satisfy the stricter EM criterion.

\section{Supplementary Results}

\subsection{Analysis of RL Training Settings}
\label{RLtraining}

\begin{table}[t]
    \centering
    
    \renewcommand{\arraystretch}{1.1}
    
    \resizebox{\columnwidth}{!}{
    \setlength{\tabcolsep}{2.8pt}
    \begin{tabular}{l @{\extracolsep{4pt}} c c c @{\hspace{10pt}} c c c}
        \toprule
        \multirow{3}{*}{\textbf{Training Setting}} & \multicolumn{6}{c}{\textbf{MMQA}} \\
        \cmidrule(lr){2-7} & \multicolumn{3}{c}{2-table} & \multicolumn{3}{c}{3-table} \\
        \cmidrule(lr){2-4}\cmidrule(lr){5-7} & EM & ECS & ES & EM & ECS & ES \\
        \midrule
        \addlinespace[2pt]
        Sequential & 44.80 & \textbf{97.74} & 45.83 & 31.68 & \textbf{92.09} & 34.78 \\[4pt]
        By-Epoch (ours) & \textbf{70.59} & 95.02 & \textbf{72.38} & \textbf{61.76} & \underline{88.24} & \underline{68.89} \\[4pt]
        By-Half-Epoch & \underline{69.23} & \underline{95.48} & \underline{71.09} & \textbf{61.76} & 87.25 & \textbf{70.79} \\[4pt]
        By-Step & 65.16 & 94.12 & 69.23 & \underline{59.80} & 
        \underline{88.24} & 67.78 \\
        \bottomrule
    \end{tabular}
    }
    \caption{Comparison of sequential and alternating dual-agent training schedules on MMQA. The By-Epoch and By-Half-Epoch variants switch the optimized agent every epoch and half-epoch, respectively. The By-Step variant first completes one epoch-level alternation cycle and then switches the active agent after every optimization step.}\label{tab:RLsetting}
\end{table}

\paragraph{Analysis.} In the Sequential setting, we first optimize the EC Agent and then optimize the RA Agent once, without returning to update the EC Agent. As shown in Table~\ref{tab:RLsetting}, all alternating schedules substantially outperform sequential training in EM and ES, although Sequential achieves the highest ECS. This indicates that one-pass optimization can produce executable programs but cannot continuously improve their downstream sufficiency: after the RA Agent adapts to the trained EC Agent, the frozen EC Agent can no longer incorporate the resulting utility feedback. Alternating optimization instead enables reciprocal co-adaptation, whereby evolving evidence distributions improve the RA Agent and its downstream feedback subsequently refines the EC Agent. We further analyze the sensitivity to the number of alternations and find that the epoch- and half-epoch schedules yield comparable results, while the By-Step schedule shows weaker performance. We therefore adopt epoch-level alternation because it achieves the highest two-table EM and ties for the highest three-table EM while maintaining competitive ECS and ES. In the By-Step setting, training first completes one round of epoch-level alternation and then switches the active agent after every optimization step. Its weaker results suggest that such high-frequency switching provides insufficient consecutive updates for either agent to adapt to a temporarily stable partner, preventing the newly acquired policy changes and utility signals from being adequately consolidated.

\subsection{Inference Latency}\label{sec:Latency}
\begin{table}[t]
    \centering
    
    \renewcommand{\arraystretch}{1.1}
    
    \resizebox{\columnwidth}{!}{
    \setlength{\tabcolsep}{2.8pt}
    \begin{tabular}{l @{\extracolsep{4pt}} *{5}{>{\centering\arraybackslash}p{1.3cm}} @{\hspace{10pt}} c}
        \toprule
        \multirow{2}{*}{\textbf{Method}} & \multicolumn{5}{c}{\textbf{MMQA}} & \multirow{2}{*}{\textbf{Thinking}} \\
        \cmidrule(lr){2-6}
         & E2E & LLM & Code & TTFT(ms) & Gen. Tokens & \\
        \midrule
        \addlinespace[2pt]
        Table-R1 & 12.67 & 12.67 & 0.000 & 31.68 & 1821 & \ding{51} \\[4pt]
        EviDARL & 5.67 & 5.66 & 0.014 & 5670.6 & 523 & \ding{55} \\[4pt]
        EC Agent & 4.93 & 4.91 & 0.014 & 24.6 & 457 & \ding{55} \\[4pt]
        RA Agent & 0.75 & 0.75 & 0.000 & 53.7 & 66 & \ding{55} \\[2pt]
        \bottomrule
    \end{tabular}
    }
    \caption{Inference latency of Table-R1 and EviDARL and the two EviDARL agent stages. E2E, LLM, and Code denote the runtime of the entire workflow, LLM inference, and code execution, respectively, measured in seconds. }
    \label{tab:InferenceLatency}
\end{table}
\paragraph{Settings.} 
We measure inference latency on four NVIDIA A100 80\,GB GPUs with 400\,GB of system memory and 48 CPU cores. Consistent with our evaluation settings, Table-R1 uses thinking mode, whereas EviDARL does not. Both methods based on \textbf{Qwen3-8B} run with the same vLLM version and configuration: \texttt{tensor\_parallel\_size=4}, \texttt{gpu\_memory\_utilization=0.85}, \texttt{max\_model\_len=16384}, BF16 precision, CUDA Graph, \texttt{seed=0}, and greedy decoding, with prefix caching disabled. Each sample is processed by an independent coroutine, with a semaphore controlling in-flight concurrency; thus, the serial overhead of EviDARL arises only from intra-sample dependencies between agents rather than cross-sample batching barriers. 
\paragraph{Analysis.} As shown in Table~\ref{tab:InferenceLatency}, EviDARL reduces end-to-end latency from 12.67\,s to 5.67\,s compared with Table-R1, corresponding to a 55.2\% reduction. Although EviDARL invokes the model multiple times and has a higher time to first token (TTFT), measured from when Agent 1 receives the input to when Agent 2 generates its first token, its agents operate without a thinking mode, thereby avoiding test-time scaling (TTS) and substantially reducing computational overhead while potentially achieving higher accuracy. This is the primary reason its overall latency remains lower than that of Table-R1. The shorter input contexts provided by compact table representations offer an additional efficiency gain. Moreover, EviDARL generates substantially fewer tokens (523 vs.\ 1,821), while code execution contributes only 0.014\,s, demonstrating that its dual-agent workflow maintains low inference cost despite the additional stage dependency.

\subsection{Ablation Study of Agent-Specific Training}
\begin{figure}[t]
    \centering
    \includegraphics[width=1.0\linewidth]{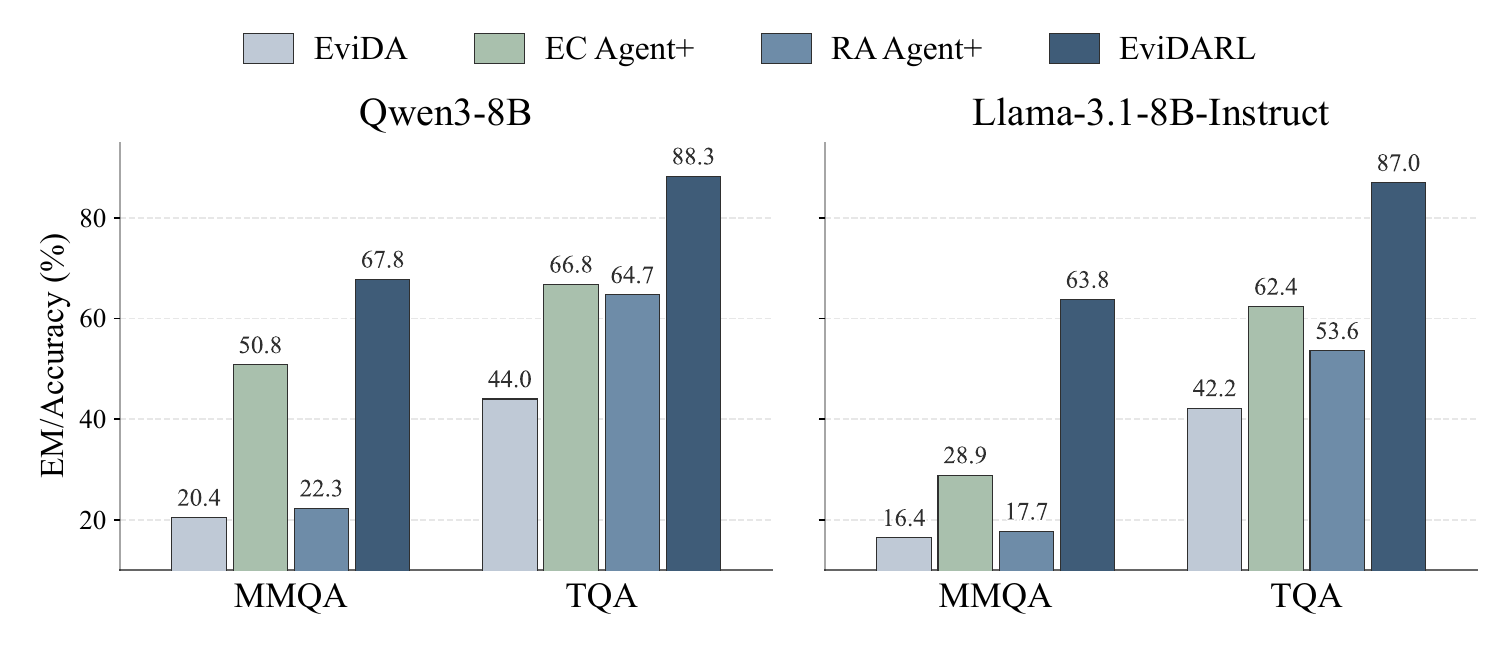}
    \caption{
    Ablation results on MMQA and TQABench with Qwen3-8B and Llama-3.1-8B-Instruct. We compare EviDARL without task-specific post-training, EC Agent+ with only the EC Agent trained, RA Agent+ with only the RA Agent trained, and EviDARL with both agents optimized through alternating RL.
    }
    \label{fig:MMQA-TQA-method-bar}
\end{figure}
\paragraph{Analysis.} MMQA reports average EM over the two- and three-table subsets, while TQABench reports average accuracy over the 8K and 64K settings. EC Agent+ denotes the variant in which only the EC Agent is optimized with RL, while the RA Agent is frozen at its base initialization (without any training). RA Agent+ is defined analogously. As shown in Figure~\ref{fig:MMQA-TQA-method-bar}, training the EC Agent yields substantially larger and more consistent gains over EviDA than training the RA Agent alone, especially on MMQA: EC Agent+ improves Qwen3-8B from 20.4 to 50.8 and Llama-3.1-8B-Instruct from 16.4 to 28.9, whereas RA Agent+ reaches only 22.3 and 17.7, respectively. This trend indicates that EviDA relies heavily on constructing accurate and sufficient evidence, which directly determines the information available for downstream reasoning. RA Agent training provides smaller gains, likely by improving evidence interpretation and downstream answer reasoning, although its effect is more visible on TQABench. EviDARL achieves the best performance in all settings, demonstrating that jointly improving evidence construction and evidence-grounded reasoning through alternating optimization produces complementary benefits. 

\end{document}